# Building a Reproducible Machine Learning Pipeline

Peter Sugimura
Tala
peter@tala.co

Florian Hartl
Tala
florian@tala.co

## ABSTRACT
Reproducibility of modeling is a problem that exists for any machine learning practitioner, whether in industry or academia. The consequences of an irreproducible model can include significant financial costs, lost time, and even loss of personal reputation (if results prove unable to be replicated). This paper will first discuss the problems we have encountered while building a variety of machine learning models, and subsequently describe the framework we built to tackle the problem of model reproducibility. The framework is comprised of four main components (data, feature, scoring, and evaluation layers), which are themselves comprised of well defined transformations. This enables us to not only exactly replicate a model, but also to reuse the transformations across different models. As a result, the platform has dramatically increased the speed of both offline and online experimentation while also ensuring model reproducibility.

## 1.INTRODUCTION
"Publish or perish" and "move fast and break things" are well-known aphorisms in academia and at startups, respectively. Both of these de-incentivize careful documentation of completed work; instead prioritizing quick release of a new product. However, reproducibility is a growing concern in the machine learning field. For example, in industry, Facebook changed their motto to "move fast with stable infrastructure". In academia, the replication crisis in psychology [3] has resulted in retracted papers. Thus far, the replication issue has been smaller in machine learning (lack of documentation as opposed to p-hacking or outright fraud in psychology), but nevertheless it is a problem. Without the ability to replicate prior results, it is difficult to compare results to determine if a newly developed algorithm is truly better than prior work [4, 7, 10].

Many open-source tools exist for the individual tasks in a machine learning pipeline. For example, Git or Subversion for version control of software code, Scikit-Learn or MLlib for building models, and Docker for containerization. However, there are few open-source software projects that link the individual tools together, which is essential for the reproducibility of models. In order to fully reproduce a machine learning pipeline, a researcher must, at a minimum, save the code, the versions of the software used, the algorithm and its hyperparameters, features and the code used to generate those features, data and the process used to obtain the data, and possibly even the specifications of the hardware used to run the model. In effect, what is needed is version control for the entire machine learning pipeline.

In the following sections, we first discuss problems we encountered while building machine learning models in production. Thereafter, we introduce our machine learning pipeline, emphasizing the pieces which enable us to ensure that we can exactly reproduce any model built using the pipeline.

## 2.PROBLEMS AND SOLUTIONS
We define reproducibility as the ability to duplicate a model exactly such that given the same raw data as input, both models return the same output. In this section, we discuss problems we faced when attempting to replicate our own results.

The root cause of most, if not all, reproducibility problems is missing information. In some cases, the existence of missing information is intentional because the underlying data is unavailable due to privacy concerns or the exact methodology is a trade secret. We will focus on the unintentional problems which hinder the ability to reproduce a model.

### 2.1.Data Provenance
Data provenance refers to the historical record of how the data of interest was collected. We have found that this is the most difficult challenge to ensure full reproducibility.

If the dataset used to train a model changes after the time of training, then it may be difficult or impossible to reproduce a model. This usually occurs in two different ways. The first is when part of the training dataset is deleted or made unavailable. The second is more subtle, and occurs if the dataset is updated. A simple example is a database table that contains an aggregate measure (i.e. count of times an event has ever occured) which is continually updated. Updated data leads to two related problems: (1) leakage due to the use of data that was not available at the time of scoring [5] and (2) concept drift due to the use of training data which is not representative of data at the time of scoring [12].

The most straightforward solution is to save a snapshot of the data every time a model is trained. However, this is impractical if the size of the dataset is large. A more practical solution is to design your data sources well, with accurate and well-documented timestamp columns. In this manner, leakage from future data is avoided by filtering out rows which contain timestamps from after the model was trained. However, in the above example of the datasource that contains the count of times an event has ever occurred, it is not possible to recover the original dataset solely from that source. In this example, one must step back and build the counts from the original data source.

A final failsafe which we have found useful is to save hashes of the content of the original training dataset. This reduces the amount of the information needed to be saved, while still ensuring that the datasets are equivalent (if the hashes match). For debugging purposes, these hashes can be saved at whatever resolution of information is necessary. For example, if the hash of the entire dataset is saved, it will be hard to diagnose where exactly in the dataset the difference originated. Finally, the method for calculating the hashes must itself also be documented and version controlled.

## 2.2. Feature Provenance

Feature provenance refers to the historical record of how a feature is generated. Any change to how a feature is generated should be tracked and version controlled. This includes other models if their predictions are used as features in a downstream model. Compared to data provenance, this problem is easier to solve since the information required is usually orders of magnitude smaller and so we can save all feature values.

To prevent unexpected changes in feature values in the first place, we follow three guidelines: (1) Individual feature generation code should be as independent from one another as possible. (2) Implemented features are immutable, i.e. in case of a bug fix for a feature, we create a new feature instead of updating the broken one. Zeldin et al. also mention this best practice [13]. (3) Because tight coupling of models can make updates difficult [9], we are very careful and deliberate when making the decision to chain or stack models.

## 2.3. Model Provenance

Model provenance refers to the record of how a model was trained. This includes the order of the features, the applied feature transformations (e.g. standardization), the hyperparameters of the algorithm, and the trained model itself. If the model is an ensemble of submodels, then the structure of the ensemble must be saved.

One easily overlooked hyperparameter is the random seed for a random number generator, which if not saved at the time the model is trained, will be impossible to recover. A more subtle error can occur if the random number generator is called more than once. In that scenario, either the random seed must be saved and set each time, or the order of transformations must be saved.

## 2.4. Software Environment

The software environment can also have an impact on reproducibility. For full reproducibility, the software versions should match exactly. For instance, even if a software package makes a bugfix after a model is trained, the original, flawed version should be used.

The solution is to use a container, or if that is not an option, then to save the versions of every software package in the environment. If this step is not done, it can be a difficult bug to identify.

## 2.5. Implementation Error

A common industry paradigm is for researchers to develop a model using their language of choice and then "throw it over the wall" to the engineering team for them to implement in a separate language. This introduces the possibility of mismatches between training and production. Thorough testing is one solution, but we choose to avoid this problem completely by using the same code for both training and production.

While this seems straightforward, we found that errors can still be introduced in two common ways. The first is human error, which can be introduced at any step in the implementation process. For instance, contents of a file might be modified or overwritten by accident. The second is deployment error ("to err is human; to really foul things up requires a computer" - Paul Ehrlich), such as when the model is unintentionally only partially deployed. For deployment errors, a reliable deployment system with detailed monitoring is worth the investment.

We have considered two solutions with respect to implementation error, both based on the idea that all information required to obtain a prediction must be linked to each other. The first method implicitly enforces these links by saving all of the information into a single unified file. However, this may not be practical for two reasons: (1) the information is heterogeneous and (2) the information may be too large to store in a single file. The second method explicitly enforces the links by calculating the hash of every file and validates that the hashes match the expected value.

## 3. SYSTEM ARCHITECTURE

We now discuss the modeling pipeline we have built, the scope of which starts from the raw data and extends to model evaluation. Key requirements of the pipeline are reproducibility, generality, scalability, and compatibility with external libraries. We ensure reproducibility by treating the concept as a first-class citizen from the start. Generality and scalability is obtained by splitting the project into components, each of which are composed of transformations. Compatibility is obtained by following Scikit-Learn's API conventions, which have become the de facto industry standard.

Scikit-Learn is an industry standard Python machine learning library with many extremely useful and powerful modules. Among them are machine learning algorithms (e.g. linear regression, random forest), preprocessing of data (e.g. missing value imputation, scaling), and model selection (e.g. feature selection, cross validation). We have used Scikit-Learn pipelines as our main inspiration for building our platform. However, we have found that Scikit-Learn, or any existing open-source machine learning package, is not sufficient to ensure full reproducibility. The reason is that, by design, the scope of these packages are limited to training a single model. As discussed in the previous section, feature provenance, data provenance, managing an ensemble of models, and saving software environments are required for full reproducibility; all of which are out of scope for existing open-source machine learning packages.

Our machine learning system architecture is illustrated in Figure 1. A central design element of the system is that all main components are used both offline as well as online. This is not only essential for assuring reproducibility but also increases reusability and speed of development. In the following we will briefly describe each layer of the architecture.

## 3.1. Data Layer

Machine learning models start with data. The data layer provides access to all of our data sources which simplifies the challenge of data provenance. It does so in the form of two distinct interfaces, both of which use SQL as the common query language under the hood. One is a general purpose interface which accepts a SQL query and a data source and makes sure that the provided data source, e.g. DynamoDB, is queried with the correct syntax and semantics. Data scientists use this interface for example for exploratory analysis. The second interface is more specific to our machine learning infrastructure. It expects a set of entities, data keys, and filter keys. Entities can be thought of as the primary keys of a database table, data keys as symbolic links to specific table columns, and filter keys describe the conditions to be used in the WHERE clause of a SQL query. Entities are themselves a combination of data keys, as are filter keys in addition to operators. The data layer generates SQL queries based on these inputs and sends them over to the general purpose interface mentioned earlier. The main consumer of the data key interface is the Feature Layer.

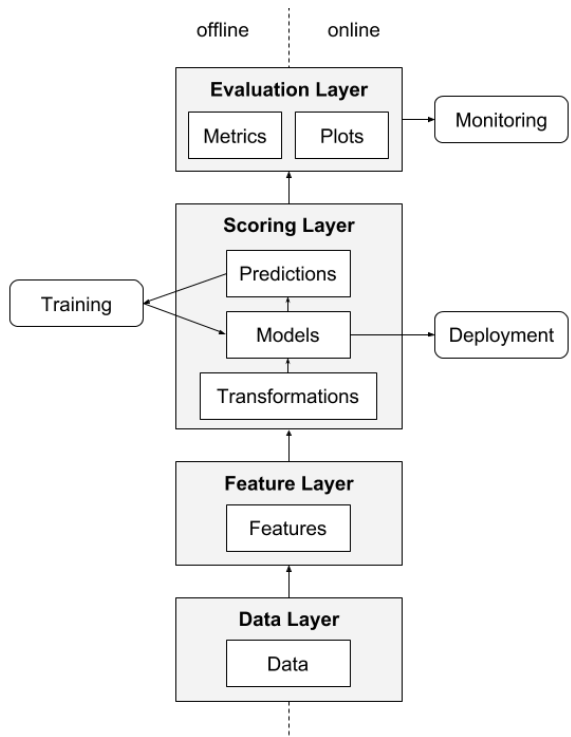

**Figure 1. Overall system architecture.**

## 3.2. Feature Layer

The feature layer is responsible for generating feature data in a transparent, reusable, and scalable manner. It is a version-controlled collection of features used throughout the company. Each feature has a unique identifier, clearly defined data requirements in form of a list of data keys, and implementation and transformation details, e.g. log-transform, assigned to it. The concept of loose coupling with the data layer through version controlled data keys ensures data provenance. As mentioned previously, we focus on independence of feature generation code without harming computational efficiency and view implemented features as immutable.

## 3.3. Scoring Layer

The scoring layer transforms features into predictions. It extends the functionality of Scikit-Learn in two ways. First, it is compatible with other machine learning libraries such as XGBoost, Vowpal Wabbit, and Tensorflow, in addition to Scikit-Learn itself. Internally, it accomplishes this by building a wrapper for each library that implements fit, transform, and predict methods. Second, it manages the workflow of an ensemble of models (Figure 2). Internally, it represents an ensemble of models as a directed acyclic graph (DAG), where each node is a model.

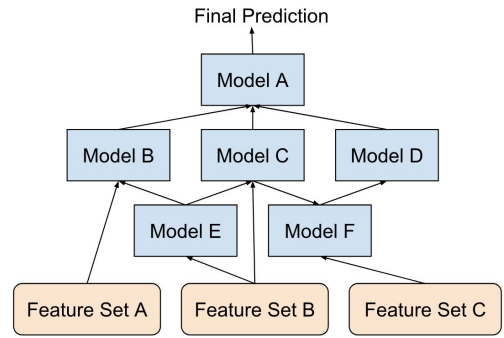

**Figure 2. An ensemble of models.**

Each model is independent from each other, but may contain common dependencies. The advantages of the scoring layer is that it can parse the DAG in order to cache repeated calculations. For example, in Figure 2, since both Model C and Model E requires the Feature Set B, the scoring layer only needs to request Feature Set B once from the Feature Layer.

Reproducibility is ensured in two ways. The first is that the offline and online scoring process uses the same set of transformations. The second, is that at the time a model is trained, everything is saved. As discussed in the prior section, this involves saving every parameter of the model; the feature provenance, data provenance, and a snapshot of the software environment. This is accomplished through the following steps:

1. Build a model config file which contains information on the features, algorithm hyperparameters, and any transformations that are applied on the input features (e.g. missing value imputation, scaling, outlier removal).
2. Save the timestamp and the random seed (based on the timestamp) to the model config file.
3. Save all information used to train the model and the trained model itself. This includes the input data, features, target labels, and software environment.
4. Save hashes of the above files to the model config file.
5. Calculate the hash of the model config file and append to its filename. This ensures that the model config file itself has not been modified.

## 3.4. Evaluation Layer

The last step in the process of ensuring model reproducibility involves the evaluation layer. This layer can check the equivalence of two models as well as evaluate the relative performance of an arbitrary number of models using a variety of model performance metrics (e.g. AUC, RMSE, log-loss). Finally, it can be used to monitor production models, to check how closely the predictions on live traffic matches the training predictions.

## 4. RELATED WORK

The issue of reproducibility in machine learning has been known for more than a decade. Sonnenburg et al. [10] call for more open-source software which we indeed have seen evolving over the past few years with Tensorflow, XGBoost, OpenML [11], and Data Version Control [1] being examples. Those software packages, however, only address part of the reproducibility problem as outlined in previous sections. Olorisade et al. [7] have identified this deficiency and include such aspects as data provenance in their list of factors affecting reproducibility of machine learning models, but they don't provide any solutions.

In recent years multiple large tech companies have built in-house machine learning platforms to increase the speed of model iteration and to make machine learning accessible to a wider range of employees. Facebook's FBLearner Flow [2] is a good example for such a system. In its essence, FBLearner Flow is a workflow scheduling and execution tool specialized to machine learning. It supports the dynamic composition of an execution pipeline consisting of operators, e.g. for splitting the dataset or training a Logistic Regression model. While our machine learning pipeline is not as flexible as FBLearner Flow due to a clear separation of responsibilities of components, we are not aware of any limitations for representing machine learning workflows with respect to our use cases. An approach which resembles ours much more is the one taken by Li et al. [6] from Uber. Their system called Michelangelo handles each step from data preparation to model serving and live monitoring in a predefined and explicit manner. They emphasize the importance of the distinction between the online and offline environments and how to solve the issues arising from the divide. Their publication, however, focuses much more on large scale data processing, including for example stream processing, and less on the conceptual details of model reproducibility or the reusability of transformation and feature extraction code. A more detailed description of the latter, including the notion of data keys, can be found in Sadekar and Jiang's work [8].

## 5.CONCLUSION AND FUTURE WORK

The process from idea to deploying a machine learning model in production includes many steps which can and should be automated and abstracted away from the machine learning practitioners to improve modeling speed and quality. An often overlooked key ingredient to realizing such benefits is reproducibility of results. We have outlined multiple difficulties of model reproducibility and described how to prevent or even solve those issues. Furthermore, we briefly presented our internal machine learning pipeline.

Future work will involve more experimentation around our data provenance approach and tooling, a more seamless mechanism to deploy models, and the addition of more functionality to the scoring and evaluation layers, e.g. to support custom machine learning models or performance evaluation techniques. We hope to open-source individual components or all of this framework in the future.

## 6.REFERENCES